\documentclass[sigconf]{acmart}
\usepackage{color}
\usepackage{algorithm2e}
\usepackage{multirow}
\usepackage{float}
\usepackage{graphicx}
\usepackage{enumitem}
\graphicspath{ {images/} }

\copyrightyear{2017}
\acmYear{2017}
\setcopyright{acmcopyright}
\acmConference{WOSP 2017}{June 19, 2017}{Toronto, ON, Canada}\acmPrice{15.00}
\acmDOI{10.1145/3127526.3127527}
\acmISBN{978-1-4503-5388-5/17/06}
\begin{document}

\title{AppTechMiner: Mining Applications and Techniques from Scientific Articles}

\author{Mayank Singh}
\authornote{First three authors have equal contribution.}
\affiliation{%
  \institution{Dept.  of Computer Science and Engg.}
  \streetaddress{IIT Kharagpur, India}
  }
\email{mayank.singh@cse.iitkgp.ernet.in}

\author{Soham Dan}
\affiliation{%
  \institution{Dept.  of Computer Science and Engg.}
  \streetaddress{IIT Kharagpur, India}
  }
\email{sohamd@cse.iitkgp.ernet.in}

\author{Sanyam Agarwal}
\affiliation{%
  \institution{Dept.  of Computer Science and Engg.}
  \streetaddress{IIT Kharagpur, India}
  }
\email{sanyama@cse.iitkgp.ernet.in}

\author{Pawan Goyal}
\affiliation{%
  \institution{Dept.  of Computer Science and Engg.}
  \streetaddress{IIT Kharagpur, India}
  }
\email{pawang@cse.iitkgp.ernet.in}

\author{Animesh Mukherjee}
\affiliation{%
  \institution{Dept.  of Computer Science and Engg.}
  \streetaddress{IIT Kharagpur, India}}
  
\email{animeshm@cse.iitkgp.ernet.in}

\renewcommand{\shortauthors}{Singh et al.}

\begin{abstract}
This paper presents \emph{AppTechMiner}, a rule-based information extraction framework that automatically constructs a knowledge base of all 
application areas and problem solving techniques. 
Techniques include tools, methods, datasets or evaluation metrics.  
We also categorize individual research articles based on their application areas and the techniques proposed/improved in the article.
Our system achieves high average precision ($\sim$82\%) and recall ($\sim$84\%) in knowledge base creation. 
It also performs well in application and technique assignment to an individual article (average accuracy $\sim$66\%).
In the end, we further present two use cases presenting a trivial information retrieval system and an extensive temporal analysis of the usage of techniques and application areas.
At present, we demonstrate the framework for the domain of computational linguistics but this can be easily generalized to any other field of research.
\end{abstract}


\maketitle

\section{Introduction}

\label{sec:intro}
It is not uncommon for researchers to envisage an information extraction system for scientific articles that can answer queries like, 
(i)\textit{What are all the techniques and tools
used in Machine Translation?}, (ii)\textit{Which are the subareas of Computational Linguistics, where Malt
Parser is frequently used?} etc. However, the meta-information necessary for constructing such a system is rarely available. Each research 
domain consists of multiple application areas which are typically associated with various techniques 
used to solve problems in these areas. 
For instance, two commonly used techniques in \textit{Information Extraction} are ``Conditional Random Fields" and 
``Hidden Markov Models". Wikipedia lists 32 popular NLP tasks and sub-tasks\footnote{https://en.wikipedia.org/wiki/Natural\_language\_processing\#Major\_tasks}. However, to our surprise, we do not find in this list many trending applications areas, for example, \textit{Dialog and Interactive systems, Social Media, Cognitive Modeling and Psycholinguistics}, etc. In addition, new techniques are continuously being proposed/improved for an application area with time and changing needs. This temporal aspect raises diverse research questions - for example,
how techniques for  \textit{POS tagging} varied over time, or, what are the most important areas of Computational
Linguistics that have been addressed in the last five years? This also should be of huge interest for new researchers surveying for an application area. 

\noindent{\bf Contributions:} In this paper, we introduce \emph{AppTechMiner} that automatically constructs a knowledge base of all application areas 
and problem solving techniques using a rule-based approach.
Subsequently, the generated knowledge base can be employed in several information retrieval systems to answer aforementioned questions. 
We demonstrate the current framework construction for the domain of computational linguistics because of the availability of full-text research articles. 
However, the proposed construction mechanism can be easily generalized to any other field of research. Next, we define two common keywords used
in the current paper:
\textbf{Area}: Area represents an application area of a particular research domain. Common application areas (hereafter written in italics) in Computational Linguistics include \textit{Machine Translation}, \textit{Dependency Parsing}, \textit{POS Tagging}, \textit{Information Extraction}, etc.\\
 
\noindent\textbf{Technique}:  A tecnhnique represents a tool or method used for a task. This may also include evaluation tool/method. Common examples (hereafter written within quotes) include ``Bleu Score", ``Rouge Score", ``Charniak Parser", ``TnT Tagger", etc. Note that technique of one paper
can potentially be an area of another paper. For example, in ``Training Nondeficient Variants of IBM-3 and IBM-4 for Word
Alignment''~\cite{schoenemann2013training}, ``Word Alignment" is an area but in ``Using Word-Dependent Transition Models in
HMM-Based Word Alignment for Statistical Machine Translation''~\cite{he2007using}, ``Word Alignment" is a technique for \textit{Machine Translation}.

The entire framework is organized into four phases (Section~\ref{method}): 
\begin{enumerate}
\item Creation of a ranked list of areas;
\item Categorizing papers on the basis of areas; 
\item Creation of a ranked list of techniques;
\item Categorizing papers on the basis of techniques.
\end{enumerate}

\noindent{\bf Key results:} We achieve high performance in each of the above phases (see Section~\ref{sec:evaluation}). 
The precision of the first phase is \textbf{84\%} (for top 30 areas) and recall is \textbf{87\%}. For the second phase, 
the accuracy is \textbf{73.3\%}. The third phase results in a precision and recall of \textbf{80\%} (for top 26 techniques) 
and \textbf{80.7\%} respectively. In the fourth phase, our system achieves an accuracy of \textbf{60\%}. 

\noindent{\bf Use cases:} In Section~\ref{use-case}, we present two use-cases: (i) constructing an information retrieval system, and 
(ii) analysis of temporal characteristics of techniques associated with an area. We also investigate the temporal variation of the popular areas 
for specific conferences, namely, \textsc{acl} and \textsc{coling}.

\section{Related Work}
Extracting application area and techniques is primarily an information extraction task. 
Information extraction (IE) from scientific articles combines approaches from natural language processing and data mining 
and has generated substantial research interest in recent times. In particular, there has been burgeoning research interest 
in the domain of biomedical documents. Shah et al.~\cite{Shah2003} extracted keywords from full text of biomedical articles and 
claim that there exist a heterogeneity in the keywords from different sections. Muller et al.~\cite{muller2008textpresso} have developed the \textit{Textpresso} framework, 
that leverage ontologies for information retrieval and extraction. In a similar work, Fukuda et al.~\cite{fukuda1998toward} proposed an IE system for protein name extraction. There has been significant work in information extraction in the area of protein structure analysis. 
Gaizauskas et al.~\cite{gaizauskas2003protein} proposed \textit{PASTA}, an IE system developed and evaluated for the protein structure domain. 
Friedman et al.~\cite{friedman2001genies} have developed a similar system to extract structure information about cellular pathways using a knowledge model. 
Biological information extraction has seen extensive work covering diverse aspects with large number of survey papers. Cohen et 
al.'s~\cite{cohen2005survey} survey on biomedical text mining, Krallinger et al.'s~\cite{krallinger2008linking} survey on 
information extraction and applications for biology and Wimalasuriya et al.~\cite{wimalasuriya2010ontology} on ontology based 
information extraction are examples of some of the popular surveys on IE for biomedical domain.

Information extraction in other domains has also received an equally strong attention from researchers. Hyponym relations have been 
extracted automatically in the celebrated work by Hearst et al.~\cite{Hearst:1992:AAH:992133.992154}. Caraballo et al.~\cite{caraballo1999automatic}
have extended previous work on automatically building semantic lexicons to automatic construction of a hierarchy of nouns and their hypernyms. 
Teufel~\cite{teufel2000argumentative} proposed information management and 
information foraging for researchers and introduced a new document analysis technique called argumentative zoning which is useful for generating user-tailored 
and task-tailored summaries. Kim et al.~\cite{kim2010semeval} and Lopez et al.~\cite{lopez2010humb} are two popular works in automatic keyphrase extraction from 
scientific articles. Quazvinian et al.~\cite{qazvinian2008scientific} have explored summarization of scientific papers using citation summary networks and citation summarization through keyphrase extraction~\cite{qazvinian2010citation}.

Jones~\cite{jones2005learning} introduced an approach for entity extraction from labeled and unlabeled text. They proposed algorithms that alternately 
look at noun phrases and their local contexts to recognize members of a semantic class in context. A relatively recent work by Gupta et
al.~\cite{gupta2014spied} developed a pattern learning system with bootstrapped entity extraction. In Gupta et al.~\cite{gupta2011analyzing}, the authors investigated the dynamics of a 
research community by extracting key aspects from scientific papers and showed how extracting key information helps in analyzing the 
influence of one community on another. 
Jin at al.~\cite{jin-EtAl:2013:EMNLP} proposed a supervised
sequence labeling system that identifies scientific terms and their accompanying definition. 

We believe that this is the first attempt to specifically mine application areas and techniques from research articles. 
Instead of complex statistical machine learning models, we employ rule-based approach, preferred in 
commercial world for information extraction tasks~\cite{DBLP:conf/emnlp/ChiticariuLR13}. 
The proposed construction mechanism can be easily generalized to any other field of research.

\section {Dataset}
We use \textsc{ACL} Anthology Network~\cite{Radev} dataset which consists of 21,213 full text papers from the domain of computational 
linguistics and natural language processing. The dataset consists of papers between the years 1965 -- 2013 from 342 \textsc{ACL} venues.

\section{Methodology}
\label{method}
In this section, we describe methods to construct knowledge base of areas and techniques.
As we already pointed out in Section~\ref{sec:intro}, the entire framework is organized in four phases: (1) creation of a ranked list of areas, (2) 
categorizing papers on the basis of areas, (3) creation of a ranked list of techniques, and (4) categorizing papers on the basis of techniques.
Next, we briefly describe these four phases in further details.

\subsection{Creation of a ranked list of areas}
\label{sec:create_rank_list}
We employ paper title information to extract areas. We use hand-written rules to extract
phrases which are likely to contain the area names. We observe that some functional keywords, such as, ``for'', ``via'', ``using'' and ``with'' act
as delimiters for such phrases. 
For example, paper title, ``Moses: Open source toolkit for statistical machine translation''~\cite{koehn2007moses} represents an instance of the 
form \textit{X for Y}, where \textit{Y} is 
the application area. We also observe that the phrase succeeding ``for'' or preceding ``using'' or both (e.g., in ``Decision procedures for 
dependency parsing using graded constraints'' \cite{menzel1998decision}) are likely to contain the name of an area. 

\noindent{\bf Seed set creation:} We create a seed set of the above functional keywords and use bootstrapped pattern learning to gather more such 
words along with areas. We had initially started with seven functional keywords and by bootstrapped pattern learning, augmented this to a final set of 11 functional keywords.

\noindent{\bf Ranking of the extracted phrases:} Even though bootstrapped pattern learning identified potential area names, 
we observe large amount of noisy phrases such as, ``machine translation system combination and evaluation''. 
Here, ``machine translation" must be extracted from the surrounding noisy words. We notice that empirical ranking algorithms produce good 
results in extraction of the exact area names from long phrases. We employ three ranking schemes, described below:
\begin{itemize}

\item \textbf{Scheme 1:} In this scheme, we rank according to individual $n$-gram scores. The score of a given $n$-gram ($N$) is calculated as:
 \begin{equation}
 Score_N = \frac{count_N}{ \sum_j{count_j}}
 \end{equation}

where, $count_N$ represents occurrence count of the $N^\textrm{th}$ $n$-gram and the denominator represents total count of all the $n$-grams. 

\item \textbf{Scheme 2:} This scheme is very similar to previous scheme with an additional constraint that if 
the score of an $n$-gram is greater than both of its border $(n-1)$ 
grams, then the border $(n-1)$ grams are left out.  The intuition behind this is as follows: the trigram ``word sense disambiguation" will have 
a higher score than its border bigrams, ``word sense'' and ``sense disambiguation'', causing both these bigrams to be left out.

\item \textbf{Scheme 3:} We improve upon the previous scheme by estimating different threshold scores for each $n$-gram. The thresholds are selected manually
by observing the individual $n$-gram lists. In Section~\ref{sec:eval_area_list}, we shall compare the precision of each of these methods 
and we have finally adopted Scheme 3 since it gives the best results.
We present 24 of the top 30 areas judged as accurate by domain experts:\end{itemize}
\fbox{%
    \parbox{0.47\textwidth}{%
        Machine Translation, Natural Language Processing, Word Sense Disambiguation, Speech Recognition, Question Answering, Dependency Parsing, Information Extraction, Chinese Word Segmentation, Semantic Role Labeling, Information Retrieval, Entity Recognition, Word Alignment, Conditional Random Fields, Maximum Entropy, Coreference Resolution, Machine Learning, Dialogue Systems, Textual Entailment, Natural Language Understanding, Active Learning, POS Tagging, Relation Extraction, Sentiment Analysis, Sense Induction
    }%
}

\subsection{Categorizing papers on the basis of areas}

In this phase, we assign individual papers to one of the discovered areas. Individual papers are categorized to their corresponding areas on the 
basis of two strategies -- direct match and relevance as per the language
models, defined for various areas.

\noindent{\bf Direct match:} In the direct match approach, we search for an explicit string match between the title or abstract and one of the areas. 
In case we do not find a match in the title, we check for a direct match with the abstract of the paper. If the abstract contains only one such 
matching area then the paper is categorized to that area. On the 
other hand, if the title or the abstract contains more than one direct match with the set of area names then we further use the language modeling approach (discussed next) to
classify that paper.\\ 
\noindent{\bf Language modeling:} In this approach, we create a language model for each area, and classify 
a document into one of these areas.
To create a language model for each area, we select the papers which could be classified on the basis of a single direct match. The titles and abstracts 
of all the papers belonging to one area are taken together to construct the language model of that area with the Jelinek-Mercer (JM) smoothing.

A document not categorized using direct match is treated as a query, consisting of the words in its title and abstract.
After experimenting on a small set of sample papers, we fixed $\lambda$ for JM smoothing to 0.7~\cite{zhai2004study}.
The prior probability $P(a)$ for an application area $a$, is proportional to the number of papers which were assigned to that
area by a single direct match of either the title or the abstract. Hence, given a query paper $q$, the area which scores the highest 

\begin{equation}
 \arg\max_a P(a|q)= \arg\max_a P(q|a)P(a)
\end{equation}

is assigned as the area for the given paper.

\subsection{Creation of a ranked list of techniques}

This extraction phase is based on the idea of {\em method} papers. We classify a paper as {\em method} paper, if it introduces a novel 
technique or provides a toolkit in an area of computational linguistics. For instance, the paper introducing the Stanford CoreNLP toolkit
is one such relevant example. 

We observe two characteristics for these papers -- one,
they are expected to have been cited a number of times which is above some threshold ($k_1$) thus indicating that the technique introduced
or improved upon is frequently used and second, the fraction of times they obtain their citations in the ``methodology'' section of other papers 
is above some threshold ($k_2$\%), thereby, indicating that they are primarily ``method papers".
In the current framework, we select $k_1 = 15$ and $k_2 = 50$\% based on extensive experiments on the AAN dataset. 
We assume that when citing paper applies a technique from the cited paper, it cites that paper and also mentions the name of the technique
in the {\em citation context} (i.e., the sentence where the citation is made). Our objective is to extract all the techniques
a method paper is used for, from the citation context(s).  We now describe the algorithm in detail.

For every method paper in the corpora, we extract all the citation contexts where this paper has been cited. We
observe that usually the techniques are represented as noun phrases in the citation contexts. For example, in the citation context, ``For English, we used the Dan Bikel implementation of the Collins parser (Collins, 2003).", we obtain three noun phrases: 1) Dan Bikel implementation, 2) Collins parser, and 3) Collins. We build a global vector of noun
phrases across all citation contexts for all the method papers. We consider this global vector as the ranked list of all the
techniques used in the computational linguistics domain. The $i$$^\textrm{th}$ component of the vector is the raw count of
the $i$$^\textrm{th}$ noun phrase, ordered lexicographically, over the method citations of the entire corpora.
Some of the top ranking noun phrases are:

\noindent\fbox{%
    \parbox{0.47\textwidth}{%
        Penn Treebank, Stanford Parser, Rate Training, Berkeley Parser, Machine Translation, Statistical Machine Translation, Charniak Parser, Moses Toolkit, Word Sense Disambiguation, Maximum Entropy, IBM Model, Bleu Score, Perceptron Algorithm, Word Alignment, Stanford POS Tagger, Collins Parser, Natural Language Processing, Bleu Metric, Coreference Resolution, Moses Decoder, Giza++ Toolkit, Brill Tagger, TnT Tagger,Anaphora Resolution, MST Parser, CCG Parser, Malt Parser, Minimum Error Rate Training
    }%
}

\subsection{Categorizing papers on the basis of techniques}
To identify the techniques for which a paper $X$ is used, we extract all the noun phrases present in all the citation context(s)
where this paper has been cited. 
We build a similar vector of these noun phrases where the $i$$^\textrm{th}$ component of the vector 
is the raw count of that noun phrase drawn from the global vector introduced in the previous section. 
If a particular noun phrase from the global vector is missing in the citation contexts for $X$, its weight is set to zero. 
We take dot product between this local vector of $X$ and the global vector to get a ranked list of possible techniques for $X$. 
Finally, we choose top $K$ techniques on this rank list as the techniques the paper $X$ is used for. 

The four phases resulted into a knowledge base that consists of a list of areas, a mapping between individual papers to the list of areas, 
a list of techniques and a mapping between individual papers to the list of techniques. We can employ this generated knowledge base in multiple 
information retrieval tasks. Section~\ref{sec:map_area_technique} demonstrate construction of one such IR system.

\section{Evaluation Results}
\label{sec:evaluation}
In this section, we present extensive evaluations carried out on our proposed system. Section~\ref{sec:eval_setup} discusses general 
evaluation guidelines along with summary of human judgment experiment settings. 

\subsection{Evaluation setup}
\label{sec:eval_setup}

As described in Section \ref{method}, the entire framework is organized into four phases. Therefore, we evaluate each phase
individually using human judgment experiments. For first and third phase, two subject experts (the first and the second author) are employed. 
For second and fourth phase, we float an online survey among six subject experts (four PhD and two under-graduate students).
Each subject expert has evaluated 20 paper-area and ten paper-technique assignments. In total, we evaluate 120 paper-area and 60 paper-technique assignments. 

\subsection{Evaluation of the ranked list of areas}
\label{sec:eval_area_list}
First, we conduct experiment to understand the relative performances of the three schemes described in Section \ref{sec:create_rank_list} for creation
of the ranked list. Scheme 3 (80\%) outperforms scheme 1 (57\%) and 2 (73\%) in terms of precision. 
Therefore, we employ scheme 3 for the creation of the ranked list in the subsequent stages.

We evaluate the ranked list of the potential areas in the computational linguistics domain extracted from the \textsc{ACL} corpora.
We employ precision-recall measures for the purpose of evaluation. 
For computing recall, however, due to limited human resource for this challenging task of labeling areas for the entire corpus of papers, we select 
a random set of 200 research papers and manually\footnote{The second author participated in labeling task.} identified each of their areas. In total, 
we find 23 distinct areas (comparable to Wikipedia list of 32 popular tasks\footnote{https://en.wikipedia.org/wiki/Natural\_language\_processing\#Major\_tasks}). Scheme 3 identified 20 out of 23 areas, achieving a high recall of 87\%.

Precision was computed by measuring fraction of correctly identified areas in the top $K$ area list. Table~\ref{tab1} presents the values of 
precision obtained for $K= 25,50,75$ and $100$ top application areas. As we can observe, majority of correct areas are ranked higher
by our ranking methodology.
\begin{table}[!tbh]
\centering
\begin{tabular}{ |c|c|c| }
  \hline
 \multirow{2}{*}{K}&\multicolumn{2}{|c|}{Precision (\%)} \\\cline{2-3}
 &Areas& Techniques\\
 \hline
  25 & 84 & 80 \\
  50 & 72 & 64\\
  75 & 51 & 48\\
  100 & 43 & 41 \\
  \hline
\end{tabular}
\caption{The precision values for $K$ = 25, 50, 75 and 100 for extraction of the list of application areas (Scheme 3) and techniques.}
\label{tab1}
\vspace{-0.5cm}
\end{table}

We also employed another domain expert to annotate first 30 results independent of the first judge. Inter-annotator agreement (Cohen's kappa coefficient) was calculated and the value of $\kappa$ came out to be \textsc{0.79}. The matrix with the agreement/disagreement count between the experts is presented in Table ~\ref{tabar}.

\begin{table}[!tbh]
 \resizebox{1\textwidth}{!}{\begin{minipage}{\textwidth}
\begin{tabular}{l|l|c|c|c|}
\multicolumn{2}{c}{}&\multicolumn{2}{c}{Domain Expert 2}&\multicolumn{1}{c}{}\\
\cline{3-5}
\multicolumn{2}{c|}{}&Yes&No&\multicolumn{1}{c|}{Total}\\
\cline{2-5}
\multirow{2}{*}{Domain Expert 1}& Yes & $23$ & $1$ & $24$\\
\cline{2-5}
& No & $1$ & $5$ & $6$\\
\cline{2-5}
\multicolumn{1}{c|}{} & \multicolumn{1}{c|}{Total} & \multicolumn{1}{|c|}{$24$} & \multicolumn{1}{|c|}{$6$} & \multicolumn{1}{|c|}{$30$}\\
\cline{2-5}
\end{tabular}
\end{minipage}}
\caption{The matrix of agreement and disagreement between two domain experts for annotation of area list.}
\label{tabar}
\vspace{-0.5cm}
\end{table}

\subsection{Evaluating the extraction of areas from individual papers}
\label{sec:eval_area_paper}

Next, we evaluate our area assignment phase. As described in Section~\ref{sec:eval_setup}, out of the 120 expert assignments, 88 (\textbf{73.3\%}) assignments were marked as correct.

\subsection{Evaluating the list of techniques}
This evaluation task is similar to the evaluation of the ranked list of application areas (see Section~\ref{sec:eval_area_list}). However, in this case, 
recall calculation is difficult if we work with the top $K$ techniques for each method paper. To simplify the task, we proceed to calculate 
recall for only the highest ranked technique for each method paper. Again, due to resource constraints, we select a small random set of 30 papers 
and aggregate all their citation contexts from the method sections of the citing papers. Annotation of this random set resulted into 26 introduced 
or improved distinct techniques. Technique extraction algorithm obtained 21 out of 26 techniques resulting in a recall of 80.7\%. 
Table~\ref{tab1} shows the precision obtained for the technique extraction algorithm for various values of $K$. As we can observe, majority of the
correct techniques are ranked higher by our ranking algorithm.

Here again we asked another domain expert to annotate the results independent of the first judge. We also calculated the inter-annotator agreement (Cohen's kappa coefficient) for the top 25 techniques and $\kappa$ came out to be \textsc{0.65}. The matrix of agreement/disagreement counts is presented in Table~\ref{tabtech}.
\begin{table}
 \resizebox{1\textwidth}{!}{\begin{minipage}{\textwidth}
\begin{tabular}{l|l|c|c|c|}
\multicolumn{2}{c}{}&\multicolumn{2}{c}{Domain Expert 2}&\multicolumn{1}{c}{}\\
\cline{3-5}
\multicolumn{2}{c|}{}&Yes&No&\multicolumn{1}{c|}{Total}\\
\cline{2-5}
\multirow{2}{*}{Domain Expert 1}& Yes & $18$ & $2$ & $20$\\
\cline{2-5}
& No & $1$ & $4$ & $5$\\
\cline{2-5}
\multicolumn{1}{c|}{} & \multicolumn{1}{|c|}{Total} & \multicolumn{1}{|c|}{$19$} & \multicolumn{1}{|c|}{$6$} & \multicolumn{1}{|c|}{$25$}\\
\cline{2-5}
\end{tabular}
\end{minipage}}
\caption{ The matrix of agreement and disagreement between two domain experts for annotation of technique list.}
\label{tabtech}
\vspace{-1cm}
\end{table}

\subsection{Evaluating the extraction of techniques from a method paper}

For this evaluation, we employ subject experts as described in Section~\ref{sec:eval_setup}. We achieve a moderate accuracy of 60\% on set of random 60 paper-techniques assignments.

\section{Use Case}

\label{use-case}
In this section, we present two use cases. In the first use case, we demonstrate construction of an example information retrieval system.
In the second use case, we analyze the evolution of the application areas and the corresponding techniques over a given time-period.

\subsection{An example information retrieval system}
\label{sec:map_area_technique}
We demonstrate the construction of an information retrieval (IR) system that takes area name as an input and outputs a list of tools and techniques.
An example of input/output of such IR system could be: \textit{Machine Translation} $\rightarrow$ ``Word Alignment", ``Gale Church Algorithm", ``Bleu Score", ``Moses Toolkit" etc.
We propose a count update based algorithm to construct this IR system. More specifically, for each 
paper $P$, we find its area and all the techniques of the method papers that it 
cites in its methodology section and append all these techniques to the list corresponding to the extracted area for this paper.

\begin{algorithm}[!tbh]
\SetAlgoLined
\KwResult{list of techniques for that area }
 initialization $ T \leftarrow \phi $\;
 \For{$P \in Corpus$}{
  $ A \leftarrow Area (P)$
  $ T \leftarrow \phi $
  $ MSet \leftarrow MethodPapersCitedBy (P)$
  \For {$M \in MSet $}{
  	$T \leftarrow T \cup Technique(M) $
    }
  $T (A) \leftarrow T(A) \cup T $
 }
 \caption{Algorithm to generate list of techniques given area name.}
 \label{algo1}
\end{algorithm}

Here, function \textit{Area(P)} returns area of a paper P. Function \textit{Technique(M)} returns techniques introduced or improved upon 
by a method paper M. Function \textit{MethodPapersCitedBy(P)} returns all the method papers cited by paper P in its methodology section. 
A simple variation of Algorithm \ref{algo1} by keeping track of the number of times a particular technique features 
in an area can potentially trace most popular techniques for an area.

In Table~\ref{tab:mapping}, we present some of the input/output examples from higher ranked areas of Computational Linguistics. 
As we see from these examples, the techniques extracted consist of sub-tasks, tools and datasets popularly used in an area. Also, it is interesting 
to observe that the extracted techniques span a wide range of time, for example, techniques like ``Collins Parser", ``Berkeley Parser", ``Charniak Parser", 
``Stanford Parser", ``MST Parser" and ``Malt Parser" are introduced in \textit{Dependency Parsing}  at substantially different time periods.

\begin{table*}[!thb]
 \resizebox{1\textwidth}{!}{\begin{minipage}{\textwidth}
 \begin{tabular}{|c|c| }
  \hline
  \textbf{Area} & \textbf{Techniques}\\
  \hline
  {\parbox[t]{4cm}{Machine Translation}} & {\parbox[t]{13cm}{Bleu Score, Rate Training, IBM Model, Word Alignment, Moses Toolkit, Inversion Transduction Grammar, Bootstrap Resampling, Translation Model, PennTreebank, Translation Quality, Language Model, Gale Church Algorithm}}  \\
  \hline
 {\parbox[t]{4cm}{Dependency Parsing}} & {\parbox[t]{13cm}{Penn Treebank, Malt Parser, Berkeley Parser, MST Parser, Charniak Parser, Collins Parser, Maximum Entropy, Nivre's Arc-Eager, Stanford Parser, Perceptron Algorithm}}\\
 \hline
 {\parbox[t]{4cm}{Multi-document Summarization}} & {\parbox[t]{13cm}{Topic Signatures, Information Extraction, Page Rank, Klsum Summarization System, Mead Summarizer, Word Sense Disambiguation, Lexical Chains, Inverse Sentence Frequency}}\\
 \hline
 {\parbox[t]{4cm}{Word Sense\\ Disambiguation}} & {\parbox[t]{13cm}{Coarse Senses, Semcor Corpus, Senseval Competitions, Cemantic Similarity, Micro Context, Maximum Entropy, Mutual Information}}\\
 \hline
 {\parbox[t]{4cm}{Sense Induction}} & {\parbox[t]{13cm}{Word Sense Disambiguation, SemEval Word Sense Induction, Chinese Whispers, Recursive Spectral Clustering, Topic Models, Graded Sense Annotation, Ontonotes Project}}\\
 \hline
 {\parbox[t]{4cm}{Opinion Mining}} & {\parbox[t]{13cm}{Sentiment Analysis, Mutual Information, Spin Model, Subjectivity Lexicon, Semantic Role Analysis, Multiclass Clasifier, Coreference Resolution, Latent Dirichlet}}\\
 \hline
 {\parbox[t]{4cm}{Chinese Word Segmentation}} & {\parbox[t]{13cm}{Entity Recognition, Conditional Random Fields, Segmentation Bakeoff, Stanford Chinese Word Segmenter, Perceptron Algorithm , Discourse Segmentation, CRF model}}\\
  \hline
\end{tabular}
\end{minipage}}
\caption{Example application areas and corresponding techniques from AAN dataset}
\label{tab:mapping}
\vspace{-0.8cm}
\end{table*}

\subsection{Temporal Analysis}

We analyze evolution of application areas and techniques over a given time-period. Below, we present three temporal scenarios.

\subsubsection{Evolution of areas}
\begin{center}
\begin{figure}[!tbh]
\includegraphics[width=0.5\textwidth]{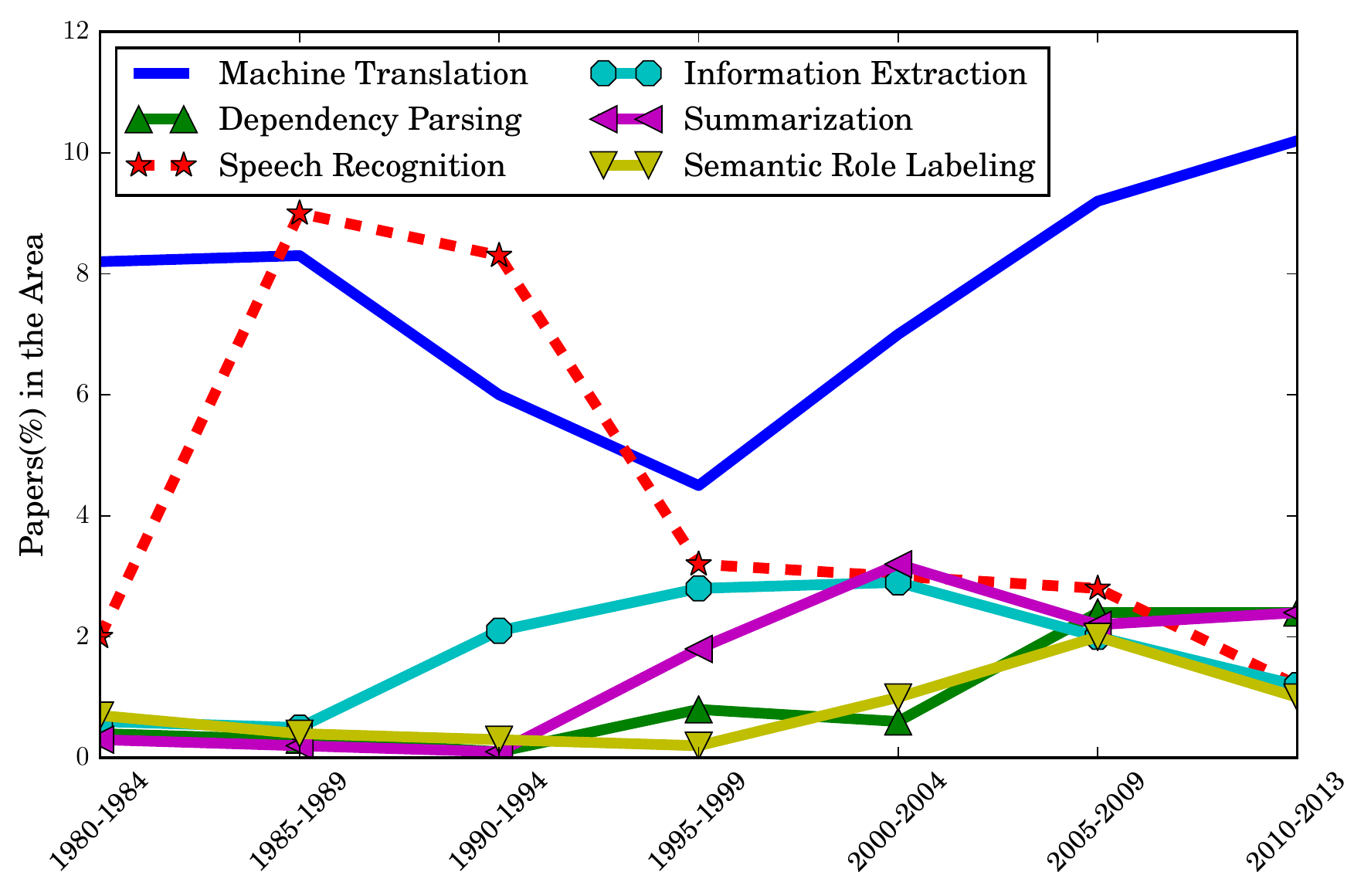}
\caption{Evolution of different application areas over time in terms of fraction of publications. \textit{Machine Translation} and \textit{dependency parsing} are on the rise, \textit{information extraction} and \textit{semantic role labeling} are on a decline. \textsc{ACL} community gradually  separates out from  \textit{Speech} community.}

\label{graph}
\vspace{-0.8cm}
\end{figure}

\end{center}

From the list of popular areas (based on the total number of papers published in an area) in \textsc{aan}, we present six representative areas, 
namely, \textit{Machine Translation}, \textit{Dependency Parsing}, \textit{Speech Recognition}, \textit{Information extraction}, \textit{Summarization} and 
\textit{Semantic Role Labeling}, and study their popularity (percentage of papers in that area for that time period out of total papers published 
in that time period) from 1980-2013 in 5-year windows. Figure~\ref{graph} demonstrates the temporal variations for these areas and how they evolve
with time. 

\noindent{\bf Observations:} While areas like  \textit{Machine Translation} and \textit{Dependency Parsing} are on the rise, 
\textit{Information extraction} and \textit{Semantic Role Labeling} are on a decline. A further interesting observation is that
till 1994, the \textsc{ACL} community had a lot of interest in \textit{Speech Recognition} which then saw a sharp decline possibly because 
of the fact that the speech community slowly separated out.

\begin{figure*}[!tbh]
\centering
 \resizebox{1\textwidth}{!}{\begin{minipage}{\textwidth}
\begin{tabular}{@{}c@{}c@{}c@{}c@{}c@{}c@{}c@{}}
  && &AAN& && \\
\includegraphics[width=.23\hsize]{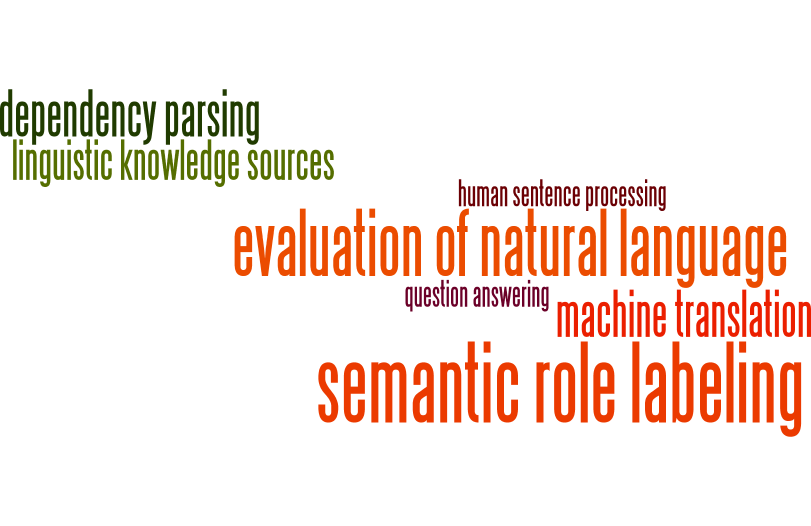} &&
 \includegraphics[width=.23\hsize]{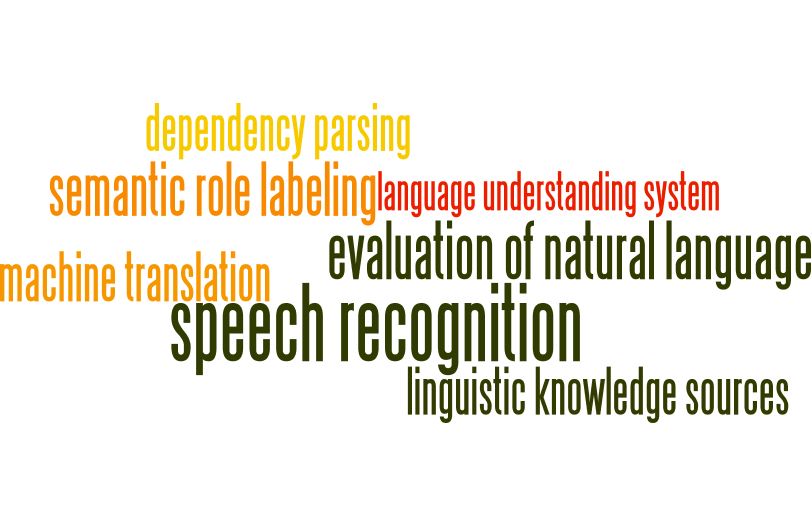} &\hspace{-20cm} &
 \includegraphics[width=.23\hsize]{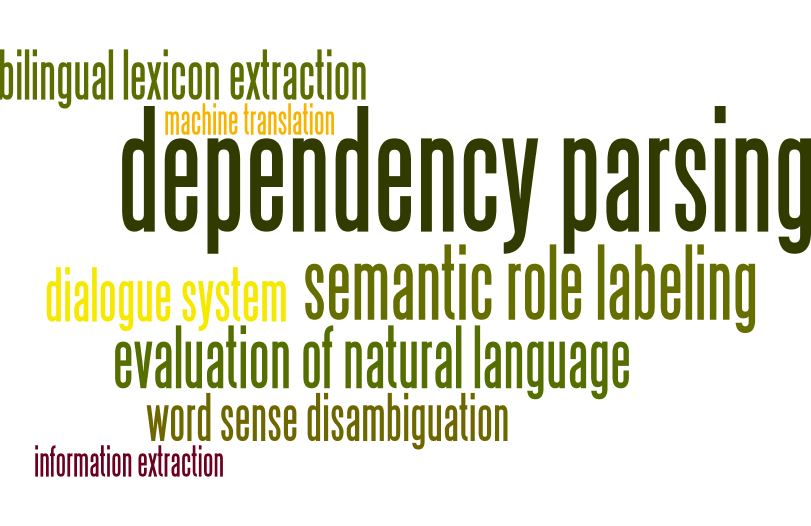} & &
 \includegraphics[width=.23\hsize]{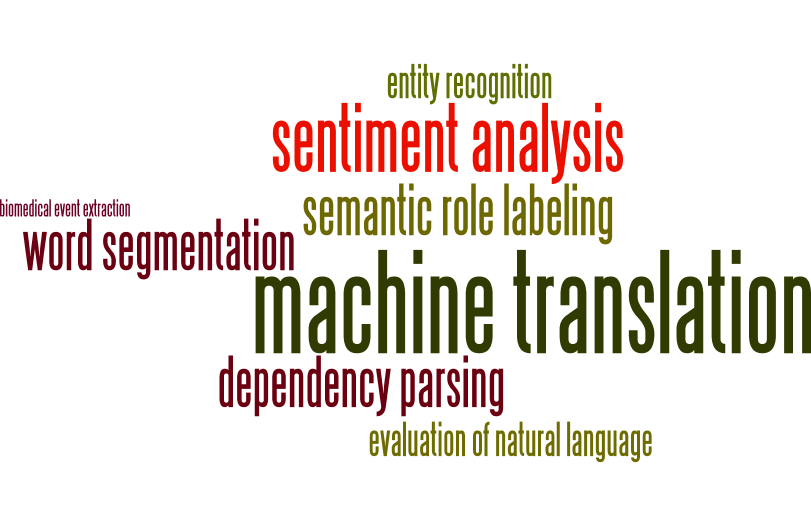} \\
  1975-1984 &&1985-1994 && 1995-2004 &&2005-2013 \\
  && && && \\
 
   && &ACL& && \\
  \includegraphics[width=.23\hsize]{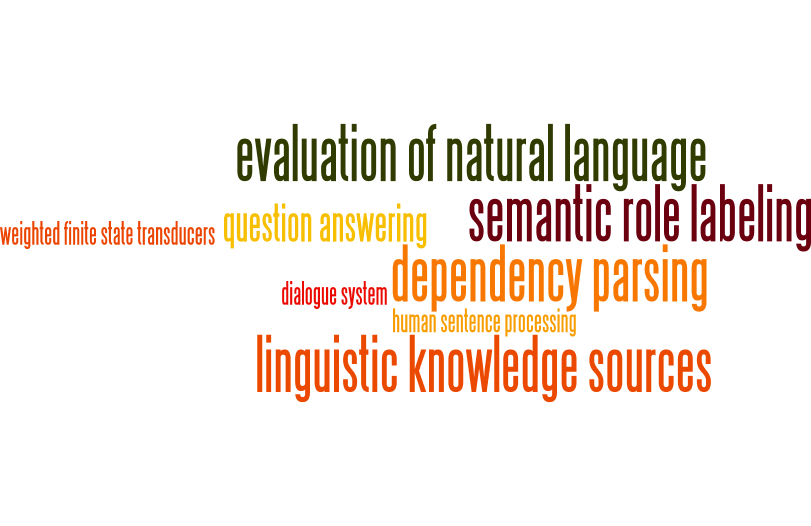} &&
  \includegraphics[width=.23\hsize]{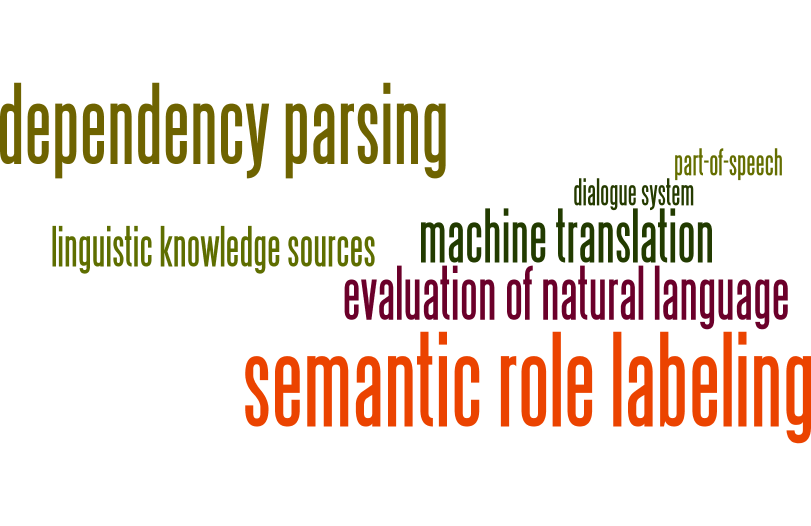} &&
  \includegraphics[width=.23\hsize]{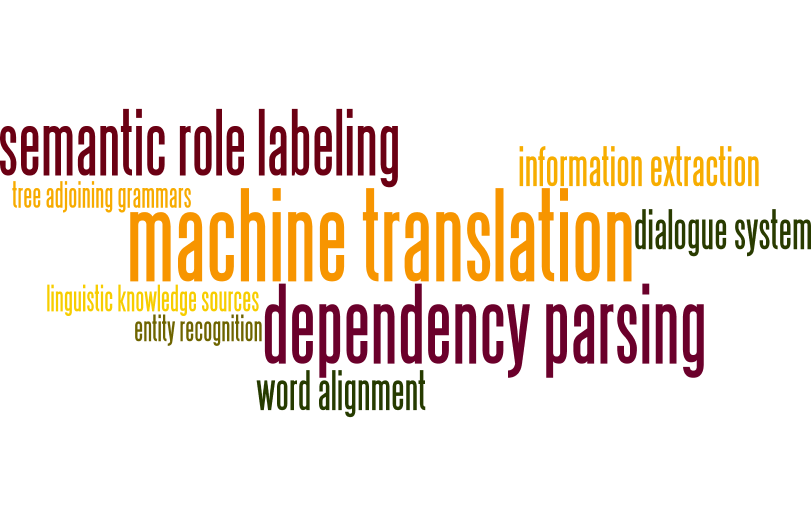} &&
  \includegraphics[width=.23\hsize]{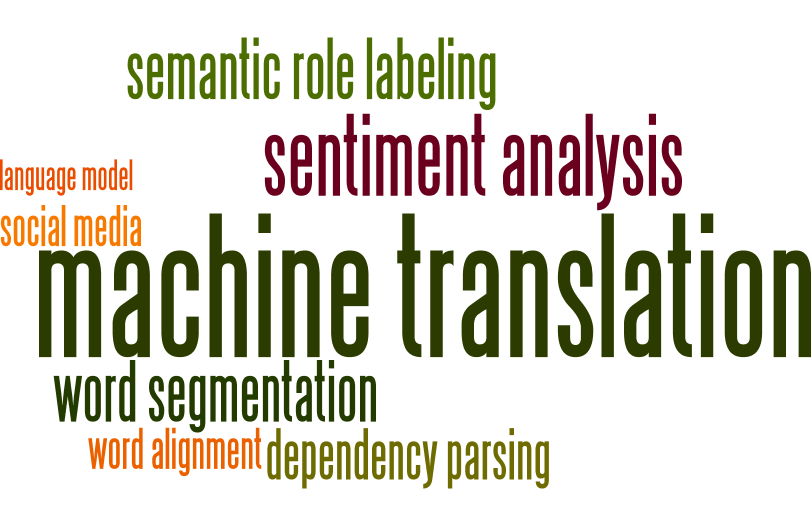} \\
  1975-1984 &&1985-1994 && 1995-2004 && 2005-2013 \\
  && && && \\

    && &COLING& &&  \\
  \includegraphics[width=.23\hsize]{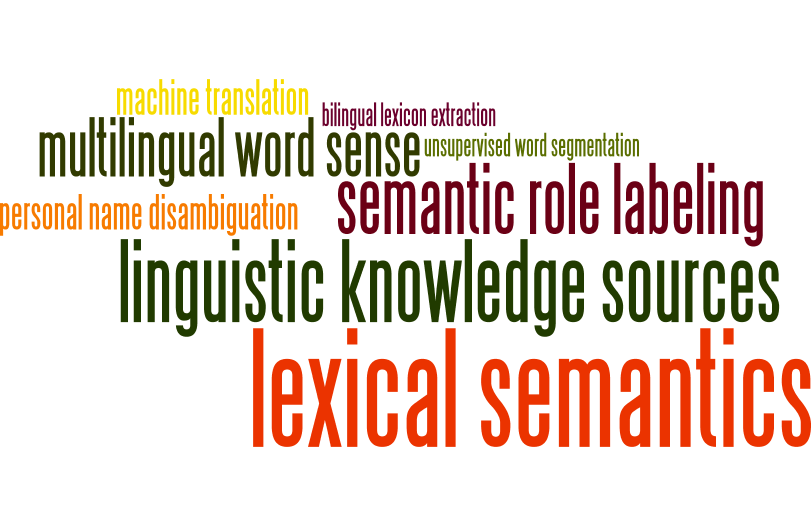} &&
  \includegraphics[width=.23\hsize]{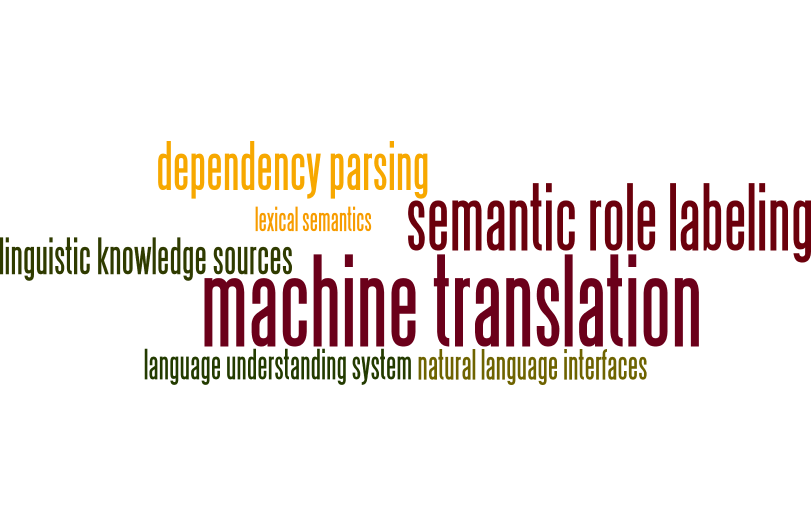} &&
   \includegraphics[width=.23\hsize]{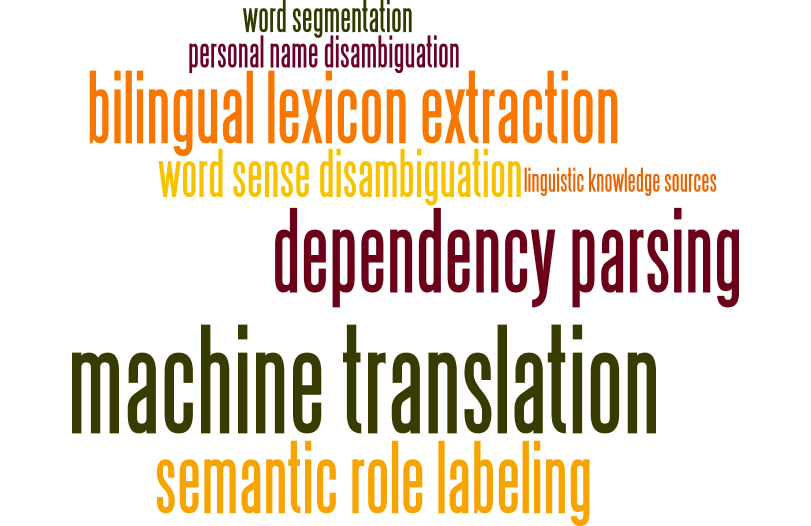} &&
  \includegraphics[width=.23\hsize]{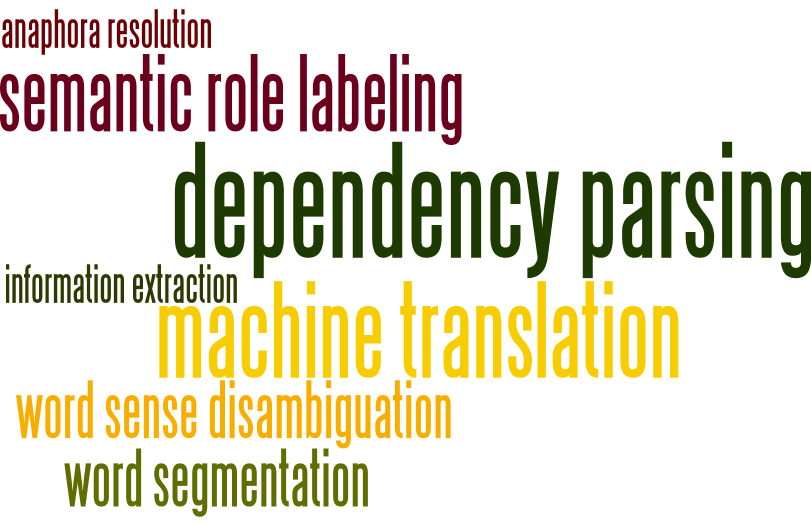} \\
  1965-1974 && 1975-1984 && 1985-1994 && 1995-2004 \\
  
\end{tabular}
\end{minipage}}
\caption{Phrase-Clouds representing the proportion of papers for an area across various time periods for the complete \textsc{AAN} dataset as well as \textsc{ACL} and \textsc{COLING} conferences. \textsc{ACL} seems to be more interested in the areas such as \textit{Machine Translation} and \textit{Dependency Parsing} over the recent decades. \textsc{COLING} community also seems more interested towards areas like \textit{Machine Translation} and \textit{Dependency Parsing}  along  with \textit{Bilingual Lexicon Extraction} in the recent decades.}  
\label{fig:word_cloud}
\vspace{-0.2cm}
\end{figure*}

\subsubsection{Evolution of major areas in top conferences}
We shortlist two top-tier conferences in the computational linguistics domain, namely, the Annual Meeting of the Association of Computational Linguistics (\textsc{ACL})
and the International Conference on Computational Linguistics (\textsc{COLING}). We study 40 years of conference history by dividing into four 10-year buckets. Next, for each conference, we extract top ten most popular areas (based on citation counts) for each 10-year bucket.
 Figure \ref{fig:word_cloud} presents phrase clouds representing evolution of areas in these two conferences in comparison to the full AAN dataset itself. 
 Some of the interesting observations from this analysis are:
 
 \begin{itemize}
\item \textbf{Full AAN dataset:} Here, we observe that while in the earlier decades, areas such as \textit{Semantic Role Labeling}, \textit{Evaluation of Natural Language} and \textit{Speech Recognition} were dominant, they fade away in the recent decades. On the other hand, areas 
such as \textit{Machine Translation} and \textit{Dependency Parsing}, which were less prevalent in the earlier decades gain significant importance in the 
recent decades. We also see \textit{Sentiment Analysis} as one of the major areas in the last decade.

\item \textbf{ACL:} In the earlier decades, this community was interested in areas like \textit{Linguistic Knowledge Sources}
and \textit{Semantic Role Labeling}. Over the recent decades, however, it seems to be more interested in areas such as \textit{Machine Translation} and
\textit{Dependency Parsing}. Interestingly, in the time period 2005 -- 2013, an upcoming area of \textit{Social Media} is found to gain importance.

\item \textbf{COLING:} Areas like \textit{Lexical Semantics} and \textit{Linguistic Knowledge Sources} were of interest to the community in the earlier decades. However, in the recent years, areas like \textit{Machine Translation}, \textit{Dependency Parsing} and \textit{Bilingual
Lexicon Extraction} have gained importance. An interesting observation here is that \textit{Semantic Role Labeling} has been all through a thrust area 
for this particular conference. 

\end{itemize}

\begin{table*}[!tbh]
\hspace{-3.2cm}
 \resizebox{0.82\textwidth}{!}{\begin{minipage}{\textwidth}
\begin{tabular}{ |c|c|c|c|c|c|}
  \hline
  \multirow{2}{*}{Area} & \multicolumn{5}{|c|}{Techniques} \\
   \cline{2-6}
  & 1990-1994&1995-1999&2000-2004&2005-2009&2010-2013\\
  \hline
{\parbox[t]{2cm}{Dependency Parsing}} & {\parbox[t]{3.5cm}{Dependency Unification Grammar, Kasper Algorithm, Left Corner Parser, Inheritence Systems, Eurotra Project}} & {\parbox[t]{3.5cm}{Penn Treebank, Probabilistic Context Free Grammar, Tree Substitution Grammar, Conditional Random Fields, Dependency Links, Collins Parser}} & {\parbox[t]{3.5cm}{Penn Treebank, Collins Parser, Berkeley Parser, Charniak Parser, Maximum Entropy, NEGRA Corpus}} & {\parbox[t]{3.5cm}{Penn Treebank, Charniak Parser, Malt Parser, MST Parser, Berkeley Parser, Stanford Parser, CCG Parser, Nivre's arc-eager}} & {\parbox[t]{3.5cm}{Penn Treebank, Malt Parser, MST Parser, Berkeley Parser, Charniak Parser, Stanford Parser, Perceptron Algorithm, Nivre's arc-eager}}\\
\hline
 {\parbox[t]{2cm}{Machine Translation}} & {\parbox[t]{3.5cm}{Parse-parse-match Approaches, Early Type Deduction, Bottom-up Head Driven Algorithm, Bilingual signs}} & {\parbox[t]{3.5cm}{IBM Model, Inversion Transduction Grammar, Word Alignment, Sentence Alignment, Moses Toolkit}} &{\parbox[t]{3.5cm}{Word Alignment, Bleu Score, Inversion Transduction Grammar, Parse-parse-match Approaches}} & {\parbox[t]{3.5cm}{Rate Training, IBM Model, Bleu Score, Word Alignment, Inversion Transduction Grammar, Moses Toolkit}} & {\parbox[t]{3.5cm}{Bleu Score, Rate Training, Moses Toolkit, Word Alignment, Bootstrap Resampling, IBM Model}}\\
\hline
{\parbox[t]{2cm}{Sentiment Analysis}} & {\parbox[t]{3.5cm}{Early Type Deduction Mechanisms, Unification Grammars, Sentence Plan Language, Mutual Information, Taxonomy Files}} & {\parbox[t]{3.5cm}{Levenshtein Distance, Discourse Structure}} & {\parbox[t]{3.5cm}{Mutual Information, Information Extraction, Penn Treebank, Distributional Similarity, Statistical Parser}} & {\parbox[t]{3.5cm}{Mutual Information, Word Sense Disambiguation, Subjectivity Lexicon, Latent Dirichlet , Spin Model}} & {\parbox[t]{3.5cm}{Mutual Information, Word Sense Disambiguation, Subjectivity Lexicon, Latent Dirichlet, Polarity Lexicons}} \\
\hline
{\parbox[t]{2cm}{Cross-lingual Textual Entailment}} & {\parbox[t]{3.5cm}{Ordinary Dictionary, Text Generation, Dependency Unification Grammar, Machine Translation}} & {\parbox[t]{3.5cm}{Discourse Structure, Encode TFS, Temporal Information, English Texts, Kappa Coefficient, CUE Phrases}} & {\parbox[t]{3.5cm}{Mutual Information, Manual Annotation, Distributional Similarity, Heuristic Approaches}} & {\parbox[t]{3.5cm}{Word Sense Disambiguation, Machine Translation, Textual Entailment Challenge}} & {\parbox[t]{3.5cm}{Semantic Textual Similarity, Verb Ocean, Moses Toolkit, Machine Translation}}\\
\hline
{\parbox[t]{2cm}{Grammatical Error Correction}} & {\parbox[t]{3.5cm}{Probabilistic Context Free Grammars, Parseval Metric, Brill POS Tagger}} & {\parbox[t]{3.5cm}{Penn Treebank, Prepositional Phrase Attachment, Collins Parser}} & {\parbox[t]{3.5cm}{Penn Treebank, Brill Tagger, FNTBL Toolkit, Charniak Parser, Kappa Statistics}} & {\parbox[t]{3.5cm}{Penn Treebank, Word Sense Disambiguation, Charniak Parser, OOV words}} & {\parbox[t]{3.5cm}{English corpus, CLC FCE dataset, OOV words, Berkeley Parser, Charniak Parser}}\\
\hline
\end{tabular}
\end{minipage}}
\caption{A few examples of areas and their top techniques for different time periods. ``Penn Treebank" is extensively used for \textit{Dependency Parsing}  and \textit{Grammatical Error Correction} across almost all time periods. ``Moses Toolkit" and ``IBM Model" are both popular techniques across most time periods in \textit{Machine Translation}. ``Mutual information" found important use in \textit{Sentiment Analysis}.}
\label{tab:technique-in-area}
\vspace{-0.7cm}
\end{table*}

\subsubsection{Evolution of techniques in areas}
In the second use case, we study evolution of techniques for a given area. For this analysis, we divide the time-line 
into fixed buckets of $4-5$ years. Next, for each bucket, we extract popular techniques (based on the number of times any paper has cited that 
technique) using our proposed system. Table~\ref{tab:technique-in-area} presents the popular techniques for five example areas. Some of  the interesting trends from Table~\ref{tab:technique-in-area} are listed below: 
\begin{itemize}\item \textbf{Dependency Parsing:} New techniques like ``Malt Parser", ``Minimum Spanning Tree (MST) Parser", etc. came into existence in 2005 -- 2009. In the next year bucket, these parsers overcome popularity of previous parsers such as ``Collin's Parser", ``Berkeley Parser" and are almost at par with ``Charniak Parser". In addition, we observe that the ``Penn Treebank" is extensively used for \textit{Dependency Parsing} across almost all time periods.
\item \textbf{Machine Translation:} We found that ``Word Alignment" and ``Inversion Transduction Grammar" are popular techniques for \textit{Machine Translation} across all time periods. Also, ``Bleu Score" has been a popular technique since its introduction in 2000 -- 2004. Similarly, ``Moses Toolkit" and ``IBM Model" are both popular techniques across most time periods.
\item \textbf{Sentiment Analysis:} In this area, ``Mutual Information" and ``Word Sense Disambiguation" are popular techniques for most of the time periods. ``Latent Dirichlet Allocation" (introduced in 2003) found important use in \textit{Sentiment Analysis} in 2005 -- 2009.  Also the ``Spin Model" got popularity in 2005 -- 2009.
\item \textbf{Cross Lingual Textual Entailment:} ``Distributional Similarity" and ``Mutual Information" are important techniques and are popular in multiple time periods. ``Verb Ocean" gets popular in 2005 -- 2009 and 2009 -- 2013. It is also very interesting to note that ``Machine Translation" is actually an important tool for this area and is very popular in 2005 -- 2009. However, in 2010 -- 2013 its popularity goes down. A probable explanation for this could be the introduction of techniques which perform \textit{Cross-lingual Textual Entailment} without ``Machine Translation"~\cite{mehdad2012fbk}.

\item \textbf{Grammatical Error Correction}: Techniques to address out-of-Vocabulary (OOV) words have become important in recent times. Over the years, ``Collins Parser" got replaced by ``Charniak Parser" and finally by ``Berkeley Parser". ``Penn Treebank" is an important dataset for this area.
\end{itemize}

\section{Conclusion and Future Work}
In this paper, we have proposed a rule-based information extraction system to extract application areas and techniques 
from scientific articles. The system extracts ranked list of all application areas in the computational linguistics domain. At a more granular level, it also extracts application area for a given paper. We evaluate our system with domain experts and prove that it performs reasonably well on both precision and recall. 
As a use case, we present an extensive analysis of temporal variation in popularity of the {\em techniques} for a given area. Some of the interesting observation that we make here are that the areas like \textit{Machine Translation} and \textit{Dependency Parsing} are on the rise of popularity while areas like \textit{Speech Recognition}, \textit{Linguistic Knowledge Sources} and \textit{Evolution of Natural Language} are on the decline.

In future, we plan to work on constructing a multi-level mapping table that maps application areas to techniques and further techniques 
to a set of parameters. For example, \textit{Machine Translation} (application area) has ``Bleu Score"
 as one of its techniques. Bleu Score is a algorithm that takes few input parameters. Changing these parameters will change the outcome of the score. Example of one such parameter is $n$, which represents the value of $n$ for the $n$-grams. 
 
All our methods can be generalized to domains other than computational linguistics. We plan to build an online version of \textit{AppTechMiner} in near future. We also plan to study temporal characteristics of techniques for a given application area to observe if future predictions can be made  for a technique - whether its popularity will increase or decrease in the years come. 

\bibliographystyle{ACM-Reference-Format}
\bibliography{sigproc} 
\end{document}